\documentclass{article}
\usepackage{ijcai22}

\usepackage{times}
\usepackage{soul}
\usepackage{url}
\usepackage[hidelinks]{hyperref}
\usepackage[utf8]{inputenc}
\usepackage[small]{caption}
\usepackage{graphicx}
\usepackage{amsmath}
\usepackage{amsthm}
\usepackage{booktabs}
\usepackage{algorithm}
\usepackage{algorithmic}

\usepackage{xspace}

\newcommand{\ie}{\emph{i.e.,}\xspace}
\newcommand{\eg}{\emph{e.g.,}\xspace}

\title{A Survey on Neural Open Information Extraction:\\ Current Status and Future Directions}

\author{
Shaowen Zhou$^1\textsuperscript{,}^2$
\and
Bowen Yu$^3$\and
Aixin Sun$^1$\and
Cheng Long$^1$\and\\
Jingyang Li$^3$\and
Haiyang Yu$^3$\and
Jian Sun$^3$\And
Yongbin Li$^3$
\affiliations
$^1$Nanyang Technological University\\
$^2$Alibaba-NTU Singapore Joint Research Institute\\
$^3$Alibaba Group, China
\emails
s200061@e.ntu.edu.sg,
yubowen.ybw@alibaba-inc.com,\\
\{axsun, c.long\}@ntu.edu.sg,
\{qiwei.ljy, yifei.yhy, jian.sun, shuide.lyb\}@alibaba-inc.com
}

\begin{document}

\maketitle

\begin{abstract}
Open Information Extraction (OpenIE) facilitates domain-independent discovery of relational facts from large corpora. The technique well suits many open-world natural language understanding scenarios, such as automatic knowledge base construction, open-domain question answering, and explicit reasoning. 
Thanks to the rapid development in deep learning technologies, numerous neural OpenIE architectures have been proposed and achieve considerable performance improvement.
In this survey, we provide an extensive overview of the state-of-the-art neural OpenIE models,  their key design decisions, strengths and weakness. 
Then, we discuss limitations of current solutions and the open issues in OpenIE problem itself. Finally we list recent trends that could help expand its scope and applicability, setting up promising directions for future research in OpenIE. To our best knowledge, this paper is the first review on neural OpenIE.
\end{abstract}

\section{Introduction}
\label{sec:intro}
Open Information Extraction (OpenIE) extracts facts in the form of $n$-ary relation tuples, \ie (\verb|arg|\textsubscript{1}, \verb|predicate|, \verb|arg|\textsubscript{2}, \ldots, \verb|arg|\textsubscript{n}), from unstructured text, without relying on predefined ontology schema \cite{niklaus-etal-2018-survey}.
Figure~\ref{fig:openieexp} shows example OpenIE tuples extracted from a given sentence. 
Compared to traditional (or closed) IE systems that request predefined relations, OpenIE relieves human labor on designing sophisticated and domain-dependent relation schema. Hence, it has the potential to handle heterogeneous corpora with minimal human intervention. 
With OpenIE, Web-scale unconstrained IE systems can be developed to acquire large quantities of knowledge.
The gathered knowledge can then be integrated and used in a wide range of natural language processing (NLP) applications, such as textual entailment \cite{berant-etal-2011-global}, summarization \cite{stanovsky-etal-2015-open}, question answering \cite{fader2014open,Mausam2016openie4}, and explicit reasoning \cite{fu-etal-2019-collaborative}.

\begin{figure}[th]
\centering
\begin{tabular}{p{8.1cm}}
    \toprule
     \textit{Deep learning is a class of ML algorithms that uses multiple layers to extract  features from the raw input.} \\
     \midrule
     (Deep learning; \textbf{is a class of}; ML algorithms)\\
     (Deep learning; \textbf{uses}; multiple layers)\\
     (Deep learning; \textbf{extracts}; features; from the raw input) \\
     \bottomrule
\end{tabular}
\caption{OpenIE tuples extracted from an example sentence (found in Wikipedia). A tuple consists of a predicate (in bold) and several arguments, representing a fact extracted from the sentence.}
\label{fig:openieexp}
\vspace{-1em}
\end{figure}

Before deep learning, traditional OpenIE systems are either statistical or rule-based, and heavily rely on the analysis of syntactic patterns \cite{niklaus-etal-2018-survey}. Recently, neural OpenIE solutions become popular, thanks to the large-scale OIE benchmarks (\eg OIE2016 \cite{stanovsky-dagan-2016-creating}, CaRB \cite{bhardwaj-etal-2019-carb}), and the great success of neural-based models on various NLP tasks (\eg NER \cite{li2020survey}, machine translation \cite{Yang2020ASO}). 
Starting with \citeauthor{stanovsky-etal-2018-supervised}~\citeyear{stanovsky-etal-2018-supervised} and \citeauthor{cui-etal-2018-neural}~\citeyear{cui-etal-2018-neural}, neural-based approaches dominate OpenIE research for their promising extraction quality on multiple OpenIE benchmarks. Neural solutions mainly formulate OpenIE as a sequence tagging problem or a sequence generation problem. 
Tagging-based methods tag a token or a span in a sentence as an argument or a predicate \cite{stanovsky-etal-2018-supervised,kolluru-etal-2020-openie6,zhanandzhao-2020-span}. 
Generative methods  generate extractions from sentence input with an auto-regressive neural architecture \cite{cui-etal-2018-neural,kolluru-etal-2020-imojie}. 
Some recent work focuses on neural model parameter calibration by introducing a new loss \cite{jiang-etal-2019-improving}, or a new objective to achieve syntactically sound and semantically consistent extraction \cite{tang-etal-2020-syntactic}.

\begin{figure*}[th]
    \centering
    \includegraphics[scale=0.7]{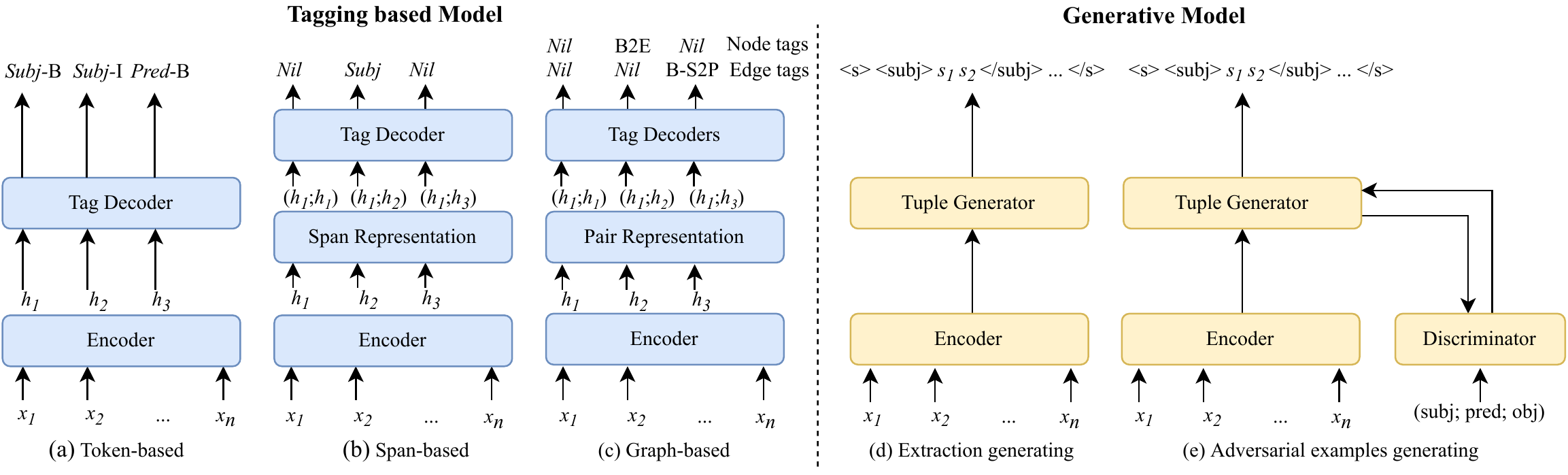}
    \caption{A taxonomy of neural OpenIE model architectures}
    \label{fig:neuralmodel}
    \vspace{-1em}
\end{figure*}

In this paper, we systematically review neural OpenIE systems.
Existing OpenIE reviews \cite{niklaus-etal-2018-survey,glauber2018systematic,claro2019multilingual} focus on traditional solutions and do not well cover the recent neural-based methods. 
Due to the paradigm change, potential avenues for future research opportunities of OpenIE need to be reconsidered as well.
In this survey, we summarise recent research developments, categorise existing neural OpenIE approaches, identify remaining issues, and discuss open problems and future directions.
The notable contributions are summarized as follows:
\textbf{1)} We propose a taxonomy of neural OpenIE models based on their task formulation. We then discuss their strengths and weaknesses;
\textbf{2)} We provide an informative discussion on the background and evaluation methods for OpenIE. We also  offer a detailed comparison of current SOTA methods; 
\textbf{3)} We discuss three challenges that restrict the development of OpenIE: evaluation, annotation, and application.  Based on them, we highlight future  directions: more open, more focused and more unified.

\section{Neural OpenIE Solutions}
\label{sec:methods}
Formally, given a sentence as a sequence of tokens/words \(S = \langle w_1, w_2, \ldots, w_n \rangle\), OpenIE outputs a list of tuples \(T={T_1, T_2, \ldots, T_p}\) with the \(i\)-th tuple \(T_i = \langle a_{i1}, p_i, a_{i2}, ..., a_{iq}\rangle \) representing a fact in the source sequence. 
Here, \(p_i\) denotes the predicate in \(T_i\), and \(a_{ij}\) is \(p_i\)'s \(j\)-th argument. The first argument in a tuple is considered as the subject. The maximum number of arguments \(m\) per tuple is pre-defined:  $m=2$ for binary and \(m \ge 3\) for $n$-ary relation extraction.

Based on task formulation, we categorize neural OpenIE models into \textit{tagging-based models} and \textit{generative models}, see Figure~\ref{fig:neuralmodel}. Next, we review architectures in the two categories, and brief solutions that focus on parameter calibration.

\subsection{Tagging-based Models}
Tagging-based models formulate OpenIE as a sequence tagging task. Given a set of tags each of which indicates a role (\eg argument, predicate) of a token or a span of tokens, the model learns the probability distribution of the tag of each token or span conditioned on sentence. Then, the OpenIE system outputs tuples based on the predicted tags.

Tagging-based OpenIE models share a similar architecture to other neural models for sequence tagging tasks in NLP (\eg NER \cite{li2020survey}). A model usually contains three modules: an \textit{embedding layer} to produce distributed representation of tokens, an \textit{encoder} to generate context-aware token representations, and a \textit{tag decoder} to predict the tag based on token representation and tagging scheme. The embedding layer often concatenates word embeddings with syntactic feature embeddings to better capture syntactic information in sentence. Recently, pre-trained language models (PLMs) have showed superior performance across various NLP tasks \cite{devlin-etal-2019-bert}. Because PLMs produce context-aware token representations, they can be used either to produce token embedding or as encoders.

Based on tagging schemes, we categorize the models into token-based, span-based, and graph-based models.

\subsubsection{Token-based Models}
Token-based models predict whether a token is (or a part of) an argument or a predicate. A common tagging scheme is BIO for \textbf{B}eginning, \textbf{I}nside, and \textbf{O}ut of a role \ie argument and predicate. Figure~\ref{fig:neuralmodel}(a) gives an example of a two-token subject and one-token predicate. A token is tagged with `O' if it is not part of an argument or predicate.

RnnOIE \cite{stanovsky-etal-2018-supervised} requires predicate head as part of input and predicts the tags indicating the arguments of the predicate. It considers part-of-speech (POS) feature, and uses Bi-directional LSTM (BiLSTM) \cite{Srivastava2015TrainingVD} to capture sentence context. It applies fully connected network with softmax layer on the output of the encoder, to produce probability distributions over all tags for each token. SenseOIE \cite{roy-etal-2019-supervising} follows RnnOIE's model structure and introduces one-hop neighbours of a token in dependency tree as syntactic features. It also adopts beam search to predict multiple relation tuples. Instead of predicting all tags in a single task, Multi\textsuperscript{2}OIE \cite{ro-etal-2020-multi} designs two sub-tasks. One predicts predicate and the other predicts the arguments that are associated to the predicted predicate. Representation of predicate tokens are used as a feature to predict arguments. The model is also the first  using PLM as sentence context encoder. OpenIE6 \cite{kolluru-etal-2020-openie6} implements an iterative grid labeling (IGL) system that organizes tag sequences in a 2D grid. Each sequence corresponds to an extraction. It uses a PLM to obtain contextualized token embedding, then feeds them to a transformer-based network \cite{vaswani-2017-attention}. The latter decodes multiple sequences of tags iteratively based on sentence input and embedding of the labels obtained in the previous step.

Token-based model is straightforward. However, arguments and predicates are often token spans. Models which predict tags for individual tokens may not well capture the span level information.

\subsubsection{Span-based Models}
Span-based models directly predict whether a \textit{token span} is an argument or a predicate.
Figure~\ref{fig:neuralmodel}(b) gives an example span \((h_1;h_2)\) which is identified as a subject from input.
Typically, all possible token spans are enumerated from input sentence. Each token span is then assigned a tag indicating its role of predicate, an argument, or otherwise not to be extracted. Enumerated token spans may overlap with others. In general, a token span representing an argument should not overlap with the one representing a predicate. This case can be handled during inference using hand-crafted constraints. 
For model design, SpanOIE~\cite{zhanandzhao-2020-span} considers POS and dependency relation between a token and its syntactic parent as syntactic features, and uses BiLSTM to produce contextualized token representation. Representation of a span is derived from the representation of its first and last tokens. Tag decoder then decodes tag from span representation.

Span-based methods consider a token span as the basic unit when deciding argument or predicate labels. This may help the model capture relationship among arguments and predicates. However, too many candidate spans that are neither argument nor predicate are generated, and it is time-consuming to enumerate all spans. Existing methods often set a maximum span length. Span-based methods also have difficulty in extracting tuple elements with discontinuous tokens, \eg \textit{``geography books''} is an argument with discontinuous tokens in sentence \textit{``Alice likes geography and history books''}. 

\subsubsection{Graph-based Models}
Graph-based models build a graph on token spans to identify triplets. MacroIE \cite{yu2021maximal} constructs a graph with nodes being token spans, and edges indicating the connected nodes belonging to the same fact. It extracts tuples by finding maximal cliques in the graph. To construct nodes, it assigns a binary indicator (\ie $B2E$ tag shown in Figure~\ref{fig:neuralmodel}(c)) to each token span; if the indicator is true, then the token span is a node. To construct edges, it assigns tags to a boundary token pair. Each token in the pair is from one token span. The assigned tag consists of two parts. The first part indicates whether the two boundary tokens are both at the beginning or at the end of the two corresponding token spans. The second part indicates the role of the two token spans. For example, $B\text{-}S2P$ tag shown in Figure~\ref{fig:neuralmodel}(c) means that token $x_1$ and $x_3$ are at the start of a subject and a predicate spans respectively.  The model learns  node and edge representations using the same architecture. It uses a BERT-based encoder to learn contextualized token representation. The model then derives span's representation from token representations, and predicts labels with a simple tag decoder.

Graph-based methods model association between tuple elements, instead of directly predicting tuples. They can extract all tuples in a single run, and better handle overlapping and discontinuous arguments or predicates. However, the current design assigns labels to all token pairs, leading to a large number of NULL labels. The imbalanced label distribution may also harm the model's performance.

\begin{table*}[th]
  \centering
  \small
  \begin{tabular}{l | c c | c c |c c | c c| c c}
    \toprule
     & \multicolumn{2}{c|}{\textbf{OIE16}}
     & \multicolumn{2}{c|}{\textbf{OIE16(S)}}
     & \multicolumn{2}{c|}{\textbf{CaRB(OIE16)}} & \multicolumn{2}{c|}{\textbf{CaRB(1-1)}} & \multicolumn{2}{c}{\textbf{CaRB}}  \\
     \textbf{OpenIE System} & F1 & AUC & F1 & AUC & F1 & AUC & F1 & AUC & F1 & AUC \\
    \midrule
    \textbf{Rule-based} & & & & & & & &  \\
    ClausIE \cite{corro-2013-clausie} & 59 & 38 & - & - & \underline{61.0} & 38.0 & 40.2 & 17.7 & 45.0 & 22.0 \\
    OpenIE4 \cite{Mausam2016openie4} & 60 & 42 & - & - & 54.3 & 37.1 & 40.5 & 20.1 & 51.6 & 29.5 \\
    \midrule
    \textbf{Tagging-based} & & & & & & & &  \\
    RnnOIE \cite{stanovsky-etal-2018-supervised} & \underline{62} & \underline{48} & 20.4 & 5.0 & 56.0 & 32.0 & 39.3 & 18.3 & 49.0 & 26.1 \\
    SenseOIE \cite{roy-etal-2019-supervising} & - & - & - & - & 31.1 & - & 23.9 & - & 28.2 & - \\
    SpanOIE \cite{zhanandzhao-2020-span} & \textbf{69.4} & \textbf{49.1} & - & - & 54.0 & - & 37.9 & - & 48.5 & - \\
    Multi\textsuperscript{2}OIE \cite{ro-etal-2020-multi} & - & - & - & - & - & - & - & - & 52.3 & 32.6 \\
    OpenIE6 \cite{kolluru-etal-2020-openie6} & - & - & - & - & \textbf{65.6} & \textbf{48.4} & 41.0 & \underline{22.9} & 52.7 & \underline{33.7} \\
    MacroIE \cite{yu2021maximal} & - & - & - & - & - & - & \textbf{43.5} & \textbf{25.0} & \textbf{54.8} & \textbf{36.3} \\
    \midrule
    \textbf{Generative} & & & & & & & & \\
    NOIE \cite{cui-etal-2018-neural} & - & 47.3 & - & - & 53.5 & 37.0 & 38.3 & 19.8 & 51.1 & 32.8 \\
    IMoJIE \cite{kolluru-etal-2020-imojie} & - & - & - & - & 56.8 & \underline{39.6} & \underline{41.2} & 22.2 &  \underline{53.3} & 33.3 \\
    \midrule
    \textbf{Calibrating RnnOIE Model} & & & & & & & & \\
    \cite{jiang-etal-2019-improving} & - & - & \underline{31.5} & \underline{12.5} & - & - & - & - & - & - \\
    \cite{tang-etal-2020-syntactic} & - & - & \textbf{32.2} & \textbf{15.9} & - & - & - & - & - & - \\
    \bottomrule
  \end{tabular}
  \caption{The performance of neural OpenIE systems on two popular benchmarks OIE2016 and CaRB, each with multiple partial matching strategies. The best results under each evaluation setting (based on the available scores) are in boldface, and the second best are underlined. The results missing in the literature are marked as ``-''. Since Logician is only evaluated on a Chinese benchmark, and Adversarial-OIE only gives precision-recall curve without AUC score on OIE2016, these two systems are not listed here. For comprehensiveness, we also include scores of two popular rule-based systems \ie ClausIE and OpenIE4.}
  \label{tab:eval}
  \vspace{-1em}
\end{table*}

\subsection{Generative Models}
Generative models formulate OpenIE as a sequence generation problem that reads a sentence and outputs a sequence of extractions.
Figure~\ref{fig:neuralmodel}(d) gives an example of the generated sequence.
Formally, given a sequence of tokens \(S\) and the expected extraction sequence \(Y = \langle y_1, y_2, \ldots, y_m \rangle\), the model maximises the conditional probability \(P(Y|S)=\prod_{i=1}^{m}{p(y_i|y_1, y_2, \ldots, y_{i-1};S)}\). There is also work which generates adversarial tuples with the goal of making it difficult for a classifier to distinguish them from  golden tuples.

\paragraph{Generate Extractions.} The generative model architecture typically consists of: \textit{an encoder} to give a distributed representation of the sentence context, and \textit{a decoder} to generate tuples sequentially, based on sentence context and the sequence generated so far. NOIE \cite{cui-etal-2018-neural} uses a 3-layer stacked LSTM as both encoder and decoder. To handle out of vocabulary (OOV) issues and retain information in source sentence, it applies a simplified copy mechanism \cite{gu-etal-2016-incorporating} to copy words from the source sentence to the generated sequences. It also applies attention mechanism \cite{bahdanau2015neural} for the RNN-based decoder to refer to the whole input sequence, instead of relying solely on the context representation produced by the encoder. Logician \cite{sun2018logician} uses bi-directional GRU \cite{cho-etal-2014-learning} as both encoder and decoder. It reduces the vocabulary size to include only predefined keywords, so that more words will be copied from the source sentence. It also implements the coverage mechanism \cite{tu-etal-2016-modeling} and explores encoding dependency parse features in the alignment model. The purpose is to reduce redundant extractions and to improve prediction accuracy. IMoJIE \cite{kolluru-etal-2020-imojie} uses BERT as encoder, and LSTM as decoder. Focusing on the redundant extraction issue in generative OpenIE models, it proposes an iterative tuple generation mechanism. This mechanism appends all tuples generated previously to the source sentence as the input, to produce the next tuple. It allows the decoder accessing all previous extractions directly, but seriously slows down the extraction speed.

\paragraph{Generate Adversarial Examples.} Adversarial-OIE \cite{han2021generative} is based on Generative Adversarial Network (GAN) \cite{Goodfellow2014GenerativeAN}. The model aims to obtain a generator which can generate tuples so similar to the gold annotations that a discriminator cannot distinguish them. The architecture consists of a transformer-based tuple generator, a Convolutional Neural Network (CNN) based discriminator, and policy gradient method REINFORCE \cite{Williams1992SimpleSG} for optimizing the generator in an adversarial manner.

\subsection{Model Comparison}
\label{sec:methods-compare}
Compared with generative models, most tagging-based models are non-autoregressive.
This fundamental difference leads to four typical model differences: 1) \textbf{Extraction dependency.} Auto-regressive models predict next tuples based on previous predictions, leading to unnecessary sequential dependency among tuples. This dependency may cause error propagation among multiple steps. At the same time, such dependency may also leverage correlation between facts, to realize reasoning for better extraction.
2) \textbf{Extraction flexibility.} Tagging-based models are not as flexible as  generative models. They assign labels to tokens and extract tokens without modification; thus the extracted tuples may be incoherent. Consider an example sentence ``\textit{Born in 1879, Albert Einstein is one of the most influential scientist of the 20th century.}'' The predicate of an extraction may be ``\textit{born in}", but a more coherent predicate is ``\textit{was born in}''. Though OpenIE6 partially solves this problem by introducing supplementary words such as \textit{``is''}, \textit{``of''} and \textit{``for''}, the cases such as predicate needs adjustment according to syntactic rules remain unsolved.
3) \textbf{Extraction faithfulness.} On the other hand, the flexibility of generative models also brings in the risk of unfaithful extraction: meaningless facts that are not expressed in the original text may be generated.
4) \textbf{Extraction speed}: Autoregressive models output results step by step. Being non-autoregressive, tagging-based methods can output results simultaneously by taking advantage of GPU parallelism. For example, the inference speed of the SOTA tagging model MacroIE~\cite{yu2021maximal} is about 35 times faster than generative model model IMoJIE~\cite{kolluru-etal-2020-imojie}.

\subsection{Calibrating Neural OpenIE Models}
Some work focuses on calibrating parameters of existing neural  models to improve extraction quality. \cite{jiang-etal-2019-improving} adds a new optimization goal to a tagging based model. The basic idea is to normalize confidence scores of the extractions, so that they are comparable across sentences. The optimization goal minimizes the binary classification loss which distinguishes correct extractions from wrong ones across different sentences. In addition, the authors also propose an iterative learning mechanism which incrementally includes extractions that participate in the computation of binary classification loss. This mechanism calibrates model parameters, and improves training examples for binary classification at the same time, leading to improved performance. \cite{tang-etal-2020-syntactic} proposes a syntactic and semantic-driven reinforcement learning method to enhance supervised OpenIE models (\eg RnnOIE). It also improves the confidence score by incorporating an extra semantic consistency score.

\section{Performance Evaluation}
\label{sec:evaluation}
In OpenIE, the input sentence is not restricted to any domain, and the extraction process does not rely on any predefined ontology schema. Hence, it becomes a challenge to derive a unified standard to judge the quality of extractions. 

Since neural-based solutions are evaluated on benchmark datasets \cite{stanovsky-dagan-2016-creating,bhardwaj-etal-2019-carb}, we first introduce their annotation standards.
Common annotation standards include completeness, correctness, and minimality. Completeness requires an OpenIE system to extract all information in a sentence. Correctness requires the extracted tuples to be implied from the sentence, and to have meaningful interpretation. Minimality requires the elements of a tuple to be indivisible units. Consider an example sentence \textit{``Jeff Bezos founded Amazon and Blue Origin".} \textit{``Amazon and Blue Origin"} should be two arguments \textit{``Amazon"} and \textit{``Blue Origin"}, \ie two extractions are formed.

\subsection{Evaluation Setting}
\label{ssec:evalsetting}
OpenIE systems are typically evaluated by comparing the extractions with the gold set. Commonly used measures are F1 and PR-AUC scores. Table \ref{tab:eval} lists the results collected from literature on two English benchmarks: OIE2016 \cite{stanovsky-dagan-2016-creating} and CaRB \cite{bhardwaj-etal-2019-carb}. 

OIE2016 is the first large-scale OpenIE benchmark. It is created by automatic conversion from QA-SRL \cite{he-etal-2015-question}, a semantic role labeling dataset. The sentences are from news (\eg WSJ) and encyclopedia (\eg WIKI) domains. Since there are no restrictions on the elements of OpenIE extractions, partial-matching criteria instead of exact-matching is typically used. Hence, the evaluation script can tolerate the extractions that are slightly different from the gold annotation. OIE2016 proposes to follow the matching criteria introduced in \cite{he-etal-2015-question}, and considers two tuples a match if both share the same grammatical head of all of the elements. However,  \cite{jiang-etal-2019-improving} noted that the evaluation metric implemented in the public code of OIE2016 uses a more lenient lexical overlap instead. \cite{jiang-etal-2019-improving} and \cite{tang-etal-2020-syntactic} follow syntactic-head matching metric and report much lower scores than those in the OIE2016 original paper. In Table~\ref{tab:eval},  columns ``OIE16'' and ``OIE16(S)'' list the results of using OIE2016 data evaluated by lexical-match and syntactic-head matching criteria respectively.

CaRB \cite{bhardwaj-etal-2019-carb} is developed by re-annotating the dev and test splits of OIE2016 via crowd-sourcing. Besides improving annotation quality, CaRB also provides a new matching scorer. CaRB scorer uses token level match and it matches relation with relation, arguments with arguments. The authors also design an extraction-gold pair match table which records the similarity scores of extraction-gold pairs for a sentence. During precision computation, each extraction is matched exclusively to one gold tuple. The extraction having the highest matching score with a gold tuple form the first exclusive match. Then the matched gold tuple is removed from the subsequent matching. The next extraction having the highest matching score with one of the remaining gold tuples forms the next exclusive match. The same matching process applies to all of the remaining extractions. Precision is the average matching scores of all extraction matches. During recall computation, CaRB scorer allows one extraction being matched by multiple gold tuples, to avoid penalizing an extraction which covers the information conveyed in multiple gold tuples. \cite{kolluru-etal-2020-openie6,yu2021maximal} also conduct experiments with other matching criteria, such as OIE2016 which is introduced earlier. They also experiment one-to-one match, which is to replace  multi-to-one mapping during recall computation with one-to-one mapping. In Table \ref{tab:eval}, the columns ``CaRB(OIE2016)'', ``CaRB(1-1)'' and ``CaRB'' list the results of using CaRB data evaluated by lexical-match, one-to-one, and the original CaRB matching criteria, respectively.

\subsection{Discussion}
\paragraph{Bootstrapping of training data.}
Training deep neural models typically requires large volume of annotated data. To obtain sufficient ``annotated data'', most neural-based OpenIE systems bootstrap training data by using existing systems (\eg rule-based systems). For example, NOIE~\cite{cui-etal-2018-neural} bootstraps  training set by applying OpenIE4~\cite{Mausam2016openie4} to a Wikipedia dump. Some work explores mixing  training samples that are produced by multiple OpenIE systems to increase sample diversity. SenseOIE \cite{roy-etal-2019-supervising} combines extractions from three OpenIE systems including Stanford OIE \cite{angeli-etal-2015-leveraging}, OpenIE5 \cite{saha-mausam-2018-open} and UKG. IMoJIE~\cite{kolluru-etal-2020-imojie} further improves the mixture quality by introducing a score-and-filter framework to denoise the extractions from multiple systems. IMoJIE reports a small increase of F1 score when compared to training using the best performing single source data. Likely, using more data sources or more advanced data argumentation techniques may further improve neural OpenIE performance. On the other hand, as the annotations are from existing systems, quality of the pseudo labels puts a limit to neural OpenIE models.

\paragraph{Common errors of neural OpenIE extractions.}
Neural OpenIE systems suffer from same common errors found in traditional systems \cite{schneider-etal-2017-analysing}. 
Besides, the limitation of neural methods also magnifies some issues. 
As discussed in \cite{kolluru-etal-2020-imojie}, generative models (\eg NOIE) suffer from redundant extractions. Due to cascading error, it is also difficult for generative models to extract all tuples when a sentence contains many gold tuples.
Extractions produced by tagging-based methods are more likely to lack auxiliary words and implied propositions. Such extractions are marked partially correct in evaluation.

\paragraph{Which model performs the best?}
We first compare results of tagging-based and generative neural OpenIE systems in Table \ref{tab:eval}. SpanOIE performs significantly better than RnnOIE on OIE16 benchmark. However, it performs slightly worse than RnnOIE on CaRB benchmark, even using the same partial-matching scorer as OIE16. This means how gold annotation is derived greatly affect the results. Without high quality benchmarks for OpenIE, it is inconclusive to state which model performs the best in general. We expect the OpenIE community to produce more benchmarks across more domains (besides news and encyclopedia), under unified annotation standard.
Another question is whether neural OpenIE systems always give higher quality extractions.
On OIE2016 benchmark, neural-based OpenIE systems achieve better F1 and AUC score than rule-based systems. However, on CaRB, rule-based OpenIE4 outperforms many neural-based OpenIE systems.
Though recent neural OpenIE systems (\eg IMoJIE, OpenIE6, and MacroIE) perform better than rule-based ones on CaRB, the improvement is not significant. To the best of our knowledge, there is no study systematically comparing neural and rule-based OpenIE systems. Note that, accuracy of current neural OpenIE systems may be limited by the low quality training data bootstrapped from rule-based systems.

\section{Challenges and Future Directions}
\label{sec:future}
OpenIE is a challenging problem due to the free form of extractions. 
Neural OpenIE systems learn high-level features automatically from training data. This new paradigm imposes new challenges and also opens up new research opportunities.
In this section, we discuss the open issues in OpenIE and set up the directions for future research.

\subsection{Challenges}
\label{sec:challenge}
\paragraph{Evaluation.}
Determining annotation specifications is difficult for OpenIE. 
Compared to closed IE which relies on predefined ontology schema in predictable domains, OpenIE imposes very few restrictions on their extractions. Thus different annotators may expect different facts to be extracted.
Due to various language phenomena in open domain, it is difficult to design a detailed and comprehensive annotation manual.
Conceptually, as long as the extracted facts are comprehensible and semantically consistent with the source text, they are considered valid extractions. 
Though recent OpenIE benchmarks provide annotation guidelines of completeness, correctness, and minimality, more detailed specifications are much expected \cite{lechelle-etal-2019-wire57}.

\paragraph{Definition.}
OpenIE is defined for open domain information extraction. However, most existing studies evaluate their solutions on news, encyclopedia, or web pages. \citeauthor{groth-etal-2018-open}~\citeyear{groth-etal-2018-open} compare  performance of traditional OpenIE systems on science, medical and general audience corpus. They find that systems perform much worse on science or medical corpus. Performance of neural OpenIE systems in domains other than news or encyclopedia is unknown, due to the lack of such benchmarks. It is also unknown how OpenIE systems perform on informal user-generated contents like tweets. Hence benchmarks covering more domains are necessary. It is also questionable whether an ominous OpenIE system that performs well on corpus in any domain is achievable. Word and grammatical patterns may vary largely in different domains.

\paragraph{Application.}
Compared to closed IE, the extractions from OpenIE are more difficult to use. There is possibility of multiple predicates referring to the same semantic relation, or arguments referring to the same entity. For example, we consider two extractions (\textit{Einstein}; \textit{was born in}; \textit{Ulm}), (\textit{Ulm}; \textit{is the birthplace of}; \textit{Einstein}). These tuples are extracted from two sentences which give the same fact. If an ontology schema is given, we may obtain a unified relation, \eg (\textit{Einstein}; \textit{schema:birthplace}; \textit{Ulm}). Moreover, recent OpenIE benchmarks (\eg CaRB) tend to keep as much relevant information as possible in gold tuples. Neural OpenIE systems optimized for these benchmarks likely output  tuples with long arguments. To remedy, recent work \cite{Wu2018TowardsPO} \cite{Vashishth2018CESICO} \cite{Pal2020CoClusteringTF} proposes to canonicalize the extracted relation tuples through clustering. \cite{gashteovski-etal-2020-aligning} explores aligning OpenIE tuples to reference knowledge bases. However, such complex remedial measures have not been fully studied. New training data, new models and new evaluation are needed, which is 
exactly a ``whack-a-mole'' situation.

\subsection{Future Directions}
\paragraph{More open.} Most existing neural OpenIE solutions follow the traditional settings that extract binary or $n$-ary tuples at sentence-level from English texts. Recently, some work explores new extraction sources either to extend the system's capability or to improve the extraction quality. New sources can be document-level texts, multilingual corpus, or multi-modal data. 
1) \textbf{Beyond sentence:} DocOIE \cite{dong-etal-2021-docoie} explores using document-level context to solve syntactic ambiguities when extracting facts at sentence-level. To facilitate further research on this topic, the authors contribute an OpenIE dataset with document-level context. Unlike the document-level relation extraction task \cite{yao-etal-2019-docred}, document-level OpenIE does not consider extracting the facts that must be inferred from more than one sentences (\eg cross-sentence co-reference). New directions may consider extracting the facts that are inferred from multiple sentences;
2) \textbf{Beyond English:} Existing neural OpenIE systems mainly focus on English corpus. The lack of multilingual OpenIE benchmarks makes it difficult to evaluate a multilingual OpenIE system's performance. To overcome this issue, a recent work \cite{ro-etal-2020-multi} attempts to use machine translation tools to create multi-language corpus, from existing English benchmarks. 
However, the performance of back translation is difficult to guarantee, which may lead to biased evaluation.
We expect high-quality human annotated benchmark to trigger more research on multilingual OpenIE.
3) \textbf{Beyond text:} Supporting extracting information from semi-structured or multi-modal data extends the capability of any extraction system including OpenIE. Openceres \cite{lockard-etal-2019-openceres} defines an OpenIE task on semi-structured websites. It utilizes the structure information to determine predicate and argument in table-like sources, though the method is not neural-based. It is also common that many web documents include images to clarify some concepts. Image itself may also contain relations. We expect future OpenIE research to explore structure and layout information in semi-structured documents, and multi-modal data.

\paragraph{More focused.} Classic OpenIE definition requires extracting all facts from the source text. 
However, in many scenarios, we are only interested in the facts that are related to certain topics/entities. The latter can be predefined.  
For example, in the task of question answering, we focus more on the facts that are related to the entities mentioned in questions, rather than all facts found in the context.
\citeauthor{yu2021semi}~\citeyear{yu2021semi} propose the concept of semi-open information extraction which restricts the subject of extractions to some entities. This definition allows OpenIE systems to focus on the facts that are directly related to predefined entities of interests. Some other work introduces more restrictions on the extraction scope and application scenarios. Assertion-based question answering (ABQA) \cite{yan2018assertion} and NeurON \cite{bhutani-etal-2019-open} extract facts from Question and Answering (QA) datasets. Here, the facts are restricted to those that answer a question. For the choices of application scenarios, ABQA targets passage-level QA data while NeurON targets conversational QA data. We expect that future work may evaluate the relatedness of extractions according to configured application scenarios, and keep those which are relevant to the application. As the result, the extractions are more focused and readily usable for downstream tasks.

\paragraph{More unified.}
OpenIE can be viewed as the most general IE task, because it includes almost all IE capabilities, such as entity recognition, relation understanding, element matching, and so on.
However, we regret to see that the IE community has not made full use of OpenIE to build a bridge between IE tasks, for a unified super IE model.
In our vision, OpenIE will become a basic pre-training objective for universal IE, leveraging its openness and generality to help a model understands what entities, relations, and facts are.

\section{Conclusion}
This survey aims to review recent progress in neural OpenIE solutions. To the best of our knowledge, we are the first to offer a  comprehensive review of the neural OpenIE solutions. We divide the neural OpenIE models into two categories:  tagging-based and generative models,  based on their task formulation. After presenting and comparing solutions in the two categories, we brief work on calibrating neural OpenIE models. In addition, we discuss the challenges of neural OpenIE solutions and outline the future directions. We hope this survey can help new researchers build a comprehensive understanding of the existing neural OpenIE solutions,  and inspire new development in this field.

\section*{Acknowledgments}
This research is supported by Alibaba Group through Alibaba Innovative Research (AIR) Program and Alibaba-NTU Singapore Joint Research Institute (JRI), and Nanyang Technological University, Singapore.

\bibliographystyle{named}
\bibliography{ijcai22}

\begin{thebibliography}{}

\bibitem[\protect\citeauthoryear{Angeli \bgroup \em et al.\egroup
  }{2015}]{angeli-etal-2015-leveraging}
Gabor Angeli, Melvin Jose~Johnson Premkumar, and Christopher~D. Manning.
\newblock Leveraging linguistic structure for open domain information
  extraction.
\newblock In {\em {ACL}}, pages 344--354, 2015.

\bibitem[\protect\citeauthoryear{Bahdanau \bgroup \em et al.\egroup
  }{2015}]{bahdanau2015neural}
Dzmitry Bahdanau, Kyunghyun Cho, and Yoshua Bengio.
\newblock Neural machine translation by jointly learning to align and
  translate.
\newblock In {\em {ICLR}}, 2015.

\bibitem[\protect\citeauthoryear{Berant \bgroup \em et al.\egroup
  }{2011}]{berant-etal-2011-global}
Jonathan Berant, Ido Dagan, and Jacob Goldberger.
\newblock Global learning of typed entailment rules.
\newblock In {\em ACL}, pages 610--619, 2011.

\bibitem[\protect\citeauthoryear{Bhardwaj \bgroup \em et al.\egroup
  }{2019}]{bhardwaj-etal-2019-carb}
Sangnie Bhardwaj, Samarth Aggarwal, and Mausam Mausam.
\newblock {C}a{RB}: A crowdsourced benchmark for open {IE}.
\newblock In {\em EMNLP-IJCNLP}, pages 6262--6267, 2019.

\bibitem[\protect\citeauthoryear{Bhutani \bgroup \em et al.\egroup
  }{2019}]{bhutani-etal-2019-open}
Nikita Bhutani, Yoshihiko Suhara, Wang-Chiew Tan, Alon Halevy, and H.~V.
  Jagadish.
\newblock Open information extraction from question-answer pairs.
\newblock In {\em NAACL}, pages 2294--2305, 2019.

\bibitem[\protect\citeauthoryear{Cho \bgroup \em et al.\egroup
  }{2014}]{cho-etal-2014-learning}
Kyunghyun Cho, Bart van Merri{\"e}nboer, Caglar Gulcehre, Dzmitry Bahdanau,
  Fethi Bougares, Holger Schwenk, and Yoshua Bengio.
\newblock Learning phrase representations using {RNN} encoder{--}decoder for
  statistical machine translation.
\newblock In {\em EMNLP}, pages 1724--1734, 2014.

\bibitem[\protect\citeauthoryear{Claro \bgroup \em et al.\egroup
  }{2019}]{claro2019multilingual}
Daniela~Barreiro Claro, Marlo Souza, Clarissa~Castell{\~{a}} Xavier, and
  Leandro~Souza de~Oliveira.
\newblock Multilingual open information extraction: Challenges and
  opportunities.
\newblock {\em Inf.}, 10(7):228, 2019.

\bibitem[\protect\citeauthoryear{Cui \bgroup \em et al.\egroup
  }{2018}]{cui-etal-2018-neural}
Lei Cui, Furu Wei, and Ming Zhou.
\newblock Neural open information extraction.
\newblock In {\em ACL}, pages 407--413, 2018.

\bibitem[\protect\citeauthoryear{Del~Corro and
  Gemulla}{2013}]{corro-2013-clausie}
Luciano Del~Corro and Rainer Gemulla.
\newblock Clausie: Clause-based open information extraction.
\newblock In {\em WWW}, page 355–366, 2013.

\bibitem[\protect\citeauthoryear{Devlin \bgroup \em et al.\egroup
  }{2019}]{devlin-etal-2019-bert}
Jacob Devlin, Ming-Wei Chang, Kenton Lee, and Kristina Toutanova.
\newblock {BERT}: Pre-training of deep bidirectional transformers for language
  understanding.
\newblock In {\em NAACL}, pages 4171--4186, 2019.

\bibitem[\protect\citeauthoryear{Dong \bgroup \em et al.\egroup
  }{2021}]{dong-etal-2021-docoie}
Kuicai Dong, Yilin Zhao, Aixin Sun, Jung{-}Jae Kim, and Xiaoli Li.
\newblock Docoie: {A} document-level context-aware dataset for openie.
\newblock In {\em {ACL/IJCNLP}}, pages 2377--2389, 2021.

\bibitem[\protect\citeauthoryear{Fader \bgroup \em et al.\egroup
  }{2014}]{fader2014open}
Anthony Fader, Luke Zettlemoyer, and Oren Etzioni.
\newblock Open question answering over curated and extracted knowledge bases.
\newblock In {\em {KDD}}, pages 1156--1165, 2014.

\bibitem[\protect\citeauthoryear{Fu \bgroup \em et al.\egroup
  }{2019}]{fu-etal-2019-collaborative}
Cong Fu, Tong Chen, Meng Qu, Woojeong Jin, and Xiang Ren.
\newblock Collaborative policy learning for open knowledge graph reasoning.
\newblock In {\em {EMNLP-IJCNLP}}, pages 2672--2681, 2019.

\bibitem[\protect\citeauthoryear{Gashteovski \bgroup \em et al.\egroup
  }{2020}]{gashteovski-etal-2020-aligning}
Kiril Gashteovski, Rainer Gemulla, Bhushan Kotnis, Sven Hertling, and Christian
  Meilicke.
\newblock On aligning {O}pen{IE} extractions with knowledge bases: A case
  study.
\newblock In {\em Proceedings of the First Workshop on Evaluation and
  Comparison of NLP Systems}, pages 143--154, 2020.

\bibitem[\protect\citeauthoryear{Glauber and
  Claro}{2018}]{glauber2018systematic}
Rafael Glauber and Daniela~Barreiro Claro.
\newblock A systematic mapping study on open information extraction.
\newblock {\em Expert Syst. Appl.}, 112:372--387, 2018.

\bibitem[\protect\citeauthoryear{Goodfellow \bgroup \em et al.\egroup
  }{2014}]{Goodfellow2014GenerativeAN}
Ian~J. Goodfellow, Jean Pouget{-}Abadie, Mehdi Mirza, Bing Xu, David
  Warde{-}Farley, Sherjil Ozair, Aaron~C. Courville, and Yoshua Bengio.
\newblock Generative adversarial nets.
\newblock In {\em NIPS}, pages 2672--2680, 2014.

\bibitem[\protect\citeauthoryear{Groth \bgroup \em et al.\egroup
  }{2018}]{groth-etal-2018-open}
Paul Groth, Michael Lauruhn, Antony Scerri, and Ron~Daniel Jr.
\newblock Open information extraction on scientific text: An evaluation.
\newblock In {\em {COLING}}, pages 3414--3423, 2018.

\bibitem[\protect\citeauthoryear{Gu \bgroup \em et al.\egroup
  }{2016}]{gu-etal-2016-incorporating}
Jiatao Gu, Zhengdong Lu, Hang Li, and Victor~O.K. Li.
\newblock Incorporating copying mechanism in sequence-to-sequence learning.
\newblock In {\em ACL}, pages 1631--1640, 2016.

\bibitem[\protect\citeauthoryear{Han and Wang}{2021}]{han2021generative}
Jiabao Han and Hongzhi Wang.
\newblock {Generative adversarial networks for open information extraction}.
\newblock {\em Advances in Computational Intelligence}, 1(4):6, 2021.

\bibitem[\protect\citeauthoryear{He \bgroup \em et al.\egroup
  }{2015}]{he-etal-2015-question}
Luheng He, Mike Lewis, and Luke Zettlemoyer.
\newblock Question-answer driven semantic role labeling: Using natural language
  to annotate natural language.
\newblock In {\em EMNLP}, pages 643--653, 2015.

\bibitem[\protect\citeauthoryear{Jiang \bgroup \em et al.\egroup
  }{2019}]{jiang-etal-2019-improving}
Zhengbao Jiang, Pengcheng Yin, and Graham Neubig.
\newblock Improving open information extraction via iterative rank-aware
  learning.
\newblock In {\em ACL}, pages 5295--5300, 2019.

\bibitem[\protect\citeauthoryear{Kolluru \bgroup \em et al.\egroup
  }{2020a}]{kolluru-etal-2020-openie6}
Keshav Kolluru, Vaibhav Adlakha, Samarth Aggarwal, {Mausam}, and Soumen
  Chakrabarti.
\newblock {O}pen{IE}6: {I}terative {G}rid {L}abeling and {C}oordination
  {A}nalysis for {O}pen {I}nformation {E}xtraction.
\newblock In {\em EMNLP}, pages 3748--3761, 2020.

\bibitem[\protect\citeauthoryear{Kolluru \bgroup \em et al.\egroup
  }{2020b}]{kolluru-etal-2020-imojie}
Keshav Kolluru, Samarth Aggarwal, Vipul Rathore, {Mausam}, and Soumen
  Chakrabarti.
\newblock {IM}o{JIE}: Iterative memory-based joint open information extraction.
\newblock In {\em ACL}, pages 5871--5886, 2020.

\bibitem[\protect\citeauthoryear{L{\'{e}}chelle \bgroup \em et al.\egroup
  }{2019}]{lechelle-etal-2019-wire57}
William L{\'{e}}chelle, Fabrizio Gotti, and Philippe Langlais.
\newblock Wire57 : {A} fine-grained benchmark for open information extraction.
\newblock In {\em Proceedings of the 13th Linguistic Annotation Workshop},
  pages 6--15, 2019.

\bibitem[\protect\citeauthoryear{Li \bgroup \em et al.\egroup
  }{2022}]{li2020survey}
Jing Li, Aixin Sun, Jianglei Han, and Chenliang Li.
\newblock A survey on deep learning for named entity recognition.
\newblock {\em {IEEE} Trans. Knowl. Data Eng.}, 34(1):50--70, 2022.

\bibitem[\protect\citeauthoryear{Lockard \bgroup \em et al.\egroup
  }{2019}]{lockard-etal-2019-openceres}
Colin Lockard, Prashant Shiralkar, and Xin~Luna Dong.
\newblock {O}pen{C}eres: {W}hen open information extraction meets the
  semi-structured web.
\newblock In {\em NAACL}, pages 3047--3056, 2019.

\bibitem[\protect\citeauthoryear{Mausam}{2016}]{Mausam2016openie4}
Mausam.
\newblock Open information extraction systems and downstream applications.
\newblock In {\em {IJCAI}}, pages 4074--4077, 2016.

\bibitem[\protect\citeauthoryear{Niklaus \bgroup \em et al.\egroup
  }{2018}]{niklaus-etal-2018-survey}
Christina Niklaus, Matthias Cetto, Andr{\'{e}} Freitas, and Siegfried
  Handschuh.
\newblock A survey on open information extraction.
\newblock In {\em {COLING}}, pages 3866--3878, 2018.

\bibitem[\protect\citeauthoryear{Pal \bgroup \em et al.\egroup
  }{2020}]{Pal2020CoClusteringTF}
Koninika Pal, Vinh~Thinh Ho, and Gerhard Weikum.
\newblock Co-clustering triples from open information extraction.
\newblock In {\em {ACM} {IKDD}}, pages 190--194, 2020.

\bibitem[\protect\citeauthoryear{Ro \bgroup \em et al.\egroup
  }{2020}]{ro-etal-2020-multi}
Youngbin Ro, Yukyung Lee, and Pilsung Kang.
\newblock {M}ulti$^2${OIE}: Multilingual open information extraction based on
  multi-head attention with {BERT}.
\newblock In {\em EMNLP}, pages 1107--1117, 2020.

\bibitem[\protect\citeauthoryear{Roy \bgroup \em et al.\egroup
  }{2019}]{roy-etal-2019-supervising}
Arpita Roy, Youngja Park, Taesung Lee, and Shimei Pan.
\newblock Supervising unsupervised open information extraction models.
\newblock In {\em EMNLP-IJCNLP}, pages 728--737, 2019.

\bibitem[\protect\citeauthoryear{Saha and
  {Mausam}}{2018}]{saha-mausam-2018-open}
Swarnadeep Saha and {Mausam}.
\newblock Open information extraction from conjunctive sentences.
\newblock In {\em COLING}, pages 2288--2299, 2018.

\bibitem[\protect\citeauthoryear{Schneider \bgroup \em et al.\egroup
  }{2017}]{schneider-etal-2017-analysing}
Rudolf Schneider, Tom Oberhauser, Tobias Klatt, Felix~A. Gers, and Alexander
  L{\"o}ser.
\newblock Analysing errors of open information extraction systems.
\newblock In {\em Proceedings of the First Workshop on Building Linguistically
  Generalizable {NLP} Systems}, pages 11--18, 2017.

\bibitem[\protect\citeauthoryear{Srivastava \bgroup \em et al.\egroup
  }{2015}]{Srivastava2015TrainingVD}
Rupesh~Kumar Srivastava, Klaus Greff, and J{\"{u}}rgen Schmidhuber.
\newblock Training very deep networks.
\newblock In {\em NIPS}, pages 2377--2385, 2015.

\bibitem[\protect\citeauthoryear{Stanovsky and
  Dagan}{2016}]{stanovsky-dagan-2016-creating}
Gabriel Stanovsky and Ido Dagan.
\newblock Creating a large benchmark for open information extraction.
\newblock In {\em EMNLP}, pages 2300--2305, 2016.

\bibitem[\protect\citeauthoryear{Stanovsky \bgroup \em et al.\egroup
  }{2015}]{stanovsky-etal-2015-open}
Gabriel Stanovsky, Ido Dagan, and {Mausam}.
\newblock Open {IE} as an intermediate structure for semantic tasks.
\newblock In {\em ACL}, pages 303--308, 2015.

\bibitem[\protect\citeauthoryear{Stanovsky \bgroup \em et al.\egroup
  }{2018}]{stanovsky-etal-2018-supervised}
Gabriel Stanovsky, Julian Michael, Luke Zettlemoyer, and Ido Dagan.
\newblock Supervised open information extraction.
\newblock In {\em NAACL}, pages 885--895, 2018.

\bibitem[\protect\citeauthoryear{Sun \bgroup \em et al.\egroup
  }{2018}]{sun2018logician}
Mingming Sun, Xu~Li, Xin Wang, Miao Fan, Yue Feng, and Ping Li.
\newblock Logician: A unified end-to-end neural approach for open-domain
  information extraction.
\newblock In {\em WSDM}, pages 556--564, 2018.

\bibitem[\protect\citeauthoryear{Tang \bgroup \em et al.\egroup
  }{2020}]{tang-etal-2020-syntactic}
Jialong Tang, Yaojie Lu, Hongyu Lin, Xianpei Han, Le~Sun, Xinyan Xiao, and Hua
  Wu.
\newblock Syntactic and semantic-driven learning for open information
  extraction.
\newblock In {\em EMNLP}, pages 782--792, 2020.

\bibitem[\protect\citeauthoryear{Tu \bgroup \em et al.\egroup
  }{2016}]{tu-etal-2016-modeling}
Zhaopeng Tu, Zhengdong Lu, Yang Liu, Xiaohua Liu, and Hang Li.
\newblock Modeling coverage for neural machine translation.
\newblock In {\em ACL}, pages 76--85, 2016.

\bibitem[\protect\citeauthoryear{Vashishth \bgroup \em et al.\egroup
  }{2018}]{Vashishth2018CESICO}
Shikhar Vashishth, Prince Jain, and Partha~P. Talukdar.
\newblock {CESI:} canonicalizing open knowledge bases using embeddings and side
  information.
\newblock In {\em WWW}, pages 1317--1327, 2018.

\bibitem[\protect\citeauthoryear{Vaswani \bgroup \em et al.\egroup
  }{2017}]{vaswani-2017-attention}
Ashish Vaswani, Noam Shazeer, Niki Parmar, Jakob Uszkoreit, Llion Jones,
  Aidan~N. Gomez, Lukasz Kaiser, and Illia Polosukhin.
\newblock Attention is all you need.
\newblock In {\em NIPS}, pages 5998--6008, 2017.

\bibitem[\protect\citeauthoryear{Williams}{1992}]{Williams1992SimpleSG}
Ronald~J. Williams.
\newblock Simple statistical gradient-following algorithms for connectionist
  reinforcement learning.
\newblock {\em Mach. Learn.}, 8:229--256, 1992.

\bibitem[\protect\citeauthoryear{Wu \bgroup \em et al.\egroup
  }{2018}]{Wu2018TowardsPO}
Tien{-}Hsuan Wu, Zhiyong Wu, Ben Kao, and Pengcheng Yin.
\newblock Towards practical open knowledge base canonicalization.
\newblock In {\em CIKM}, pages 883--892, 2018.

\bibitem[\protect\citeauthoryear{Yan \bgroup \em et al.\egroup
  }{2018}]{yan2018assertion}
Zhao Yan, Duyu Tang, Nan Duan, Shujie Liu, Wendi Wang, Daxin Jiang, Ming Zhou,
  and Zhoujun Li.
\newblock Assertion-based {QA} with question-aware open information extraction.
\newblock In {\em {AAAI}}, pages 6021--6028, 2018.

\bibitem[\protect\citeauthoryear{Yang \bgroup \em et al.\egroup
  }{2020}]{Yang2020ASO}
Shuoheng Yang, Yuxin Wang, and Xiaowen Chu.
\newblock A survey of deep learning techniques for neural machine translation.
\newblock {\em CoRR}, abs/2002.07526, 2020.

\bibitem[\protect\citeauthoryear{Yao \bgroup \em et al.\egroup
  }{2019}]{yao-etal-2019-docred}
Yuan Yao, Deming Ye, Peng Li, Xu~Han, Yankai Lin, Zhenghao Liu, Zhiyuan Liu,
  Lixin Huang, Jie Zhou, and Maosong Sun.
\newblock {D}oc{RED}: A large-scale document-level relation extraction dataset.
\newblock In {\em ACL}, pages 764--777, 2019.

\bibitem[\protect\citeauthoryear{Yu \bgroup \em et al.\egroup
  }{2021a}]{yu2021maximal}
Bowen Yu, Yucheng Wang, Tingwen Liu, Hongsong Zhu, Limin Sun, and Bin Wang.
\newblock Maximal clique based non-autoregressive open information extraction.
\newblock In {\em EMNLP}, pages 9696--9706, 2021.

\bibitem[\protect\citeauthoryear{Yu \bgroup \em et al.\egroup
  }{2021b}]{yu2021semi}
Bowen Yu, Zhenyu Zhang, Jiawei Sheng, Tingwen Liu, Yubin Wang, Yucheng Wang,
  and Bin Wang.
\newblock Semi-open information extraction.
\newblock In {\em {WWW}}, pages 1661--1672, 2021.

\bibitem[\protect\citeauthoryear{Zhan and Zhao}{2020}]{zhanandzhao-2020-span}
Junlang Zhan and Hai Zhao.
\newblock Span model for open information extraction on accurate corpus.
\newblock In {\em {AAAI}}, pages 9523--9530, 2020.

\end{thebibliography}

\end{document}